\documentclass[pdflatex,sn-mathphys-num, iicol]{sn-jnl}

\usepackage{graphicx}%
\usepackage{multirow}%
\usepackage{amsmath,amssymb,amsfonts}%
\usepackage{amsthm}%
\usepackage{mathrsfs}%
\usepackage[title]{appendix}%
\usepackage{xcolor}%
\usepackage{textcomp}%
\usepackage{manyfoot}%
\usepackage{booktabs}%
\usepackage{algorithm}%
\usepackage{algorithmicx}%
\usepackage{algpseudocode}%
\usepackage{listings}%

\usepackage{booktabs} 
\usepackage{pifont}   
\usepackage{amssymb}  
\newcommand{\xmark}{\ding{55}} 
\newcommand{\cmark}{\ding{51}}

\theoremstyle{thmstyleone}%
\theoremstyle{thmstyletwo}%

\theoremstyle{thmstylethree}%

\newcommand{\improvement}[1]{\textcolor{green!50!black}{\textsubscript{\textbf{#1}}}}

\begin{document}

\title[Article Title]{Classifying Novel 3D-Printed Objects without Retraining: Towards Post-Production Automation in Additive Manufacturing
}


\author*[1]{\fnm{Fanis} \sur{Mathioulakis}}\email{fanis.mathioulakis@kuleuven.be}

\author[1]{\fnm{Gorjan} \sur{Radevski}}

\author[2]{\fnm{Silke GC} \sur{Cleuren}}

\author[2]{\fnm{Michel} \sur{Janssens}}

\author[2]{\fnm{Brecht} \sur{Das}}

\author[3]{\fnm{Koen} \sur{Schauwaert}}

\author[1]{\fnm{Tinne} \sur{Tuytelaars}}


\affil[1]{\orgname{KU Leuven}, \country{Belgium}}

\affil[2]{\orgname{Materialise}, \country{Belgium}}

\affil[3]{\orgname{Iristick}, \country{Belgium}}



\abstract{Reliable classification of 3D-printed objects is essential for automating post-production workflows in industrial additive manufacturing. Despite extensive automation in other stages of the printing pipeline, this task still relies heavily on manual inspection, as the set of objects to be classified can change daily, making frequent model retraining impractical. Automating the identification step is therefore critical for improving operational efficiency. A vision model that could classify any set of objects by utilizing their corresponding CAD models and avoiding retraining would be highly beneficial in this setting. To enable systematic evaluation of vision models on this task, we introduce ThingiPrint, a new publicly available dataset that pairs CAD models with real photographs of their 3D-printed counterparts. Using ThingiPrint, we benchmark a range of existing vision models on the task of 3D-printed object classification. We additionally show that contrastive fine-tuning with a rotation-invariant objective allows effective prototype-based classification of previously unseen 3D-printed objects. By relying solely on the available CAD models, this avoids the need for retraining when new objects are introduced. Experiments show that this approach outperforms standard pretrained baselines, suggesting improved generalization and practical relevance for real-world use. Our dataset is publicly available at \url{https://huggingface.co/datasets/fanismathioulakis/thingiprint}.}



\maketitle

\section{Introduction}
\label{sec:intro}

The ability to accurately classify 3D-printed objects is a critical yet challenging requirement for automating and streamlining post-production workflows in industrial additive manufacturing. Despite the high degree of automation in the printing process itself, the subsequent handling of physical parts remains largely manual, representing a persistent bottleneck in production efficiency.

In high-volume production environments, multiple objects are often printed together in the same build job. After printing, these parts are typically placed into a shared bin for post-processing, at which point the association between the printed objects and their digital identities is effectively lost. In this stage, a human operator is responsible for sorting the printed objects and identifying each one from a list of possible classes, making the process laborious and time-consuming. Automating this classification step using a computer vision model can substantially improve workflow efficiency and address a real need in modern additive manufacturing environments.

This study addresses a real-world operational application in which visual identification is performed by human operators using wearable vision technology. In the considered workflow, a technician equipped with smart glasses verifies the identity of each object as it is removed from a bin. The technician manually grasps the object and captures an image in real time using the smart glasses. A vision model then performs object classification from the captured image, providing identification support during the operational process.

This operational scenario introduces two key technical challenges. (1) \textit{Scalability and generalization:} The set of printed objects changes frequently, often on a daily basis, requiring the system to recognize novel object classes without retraining. Collecting and annotating real photographs for each new print job is infeasible, and retraining the model on synthetic renderings for every new object would be highly impractical. Therefore, a scalable solution must effectively leverage synthetic images generated from available CAD models at inference while avoiding model retraining. (2) \textit{Viewpoint invariance:} Because the user manipulates and removes the object freely, the images captured by the smart glasses can exhibit viewpoint variation. The model must therefore be robust to such variations to ensure reliable classification performance.

\begin{figure}[t]
    \centering
    \includegraphics[width=\columnwidth]{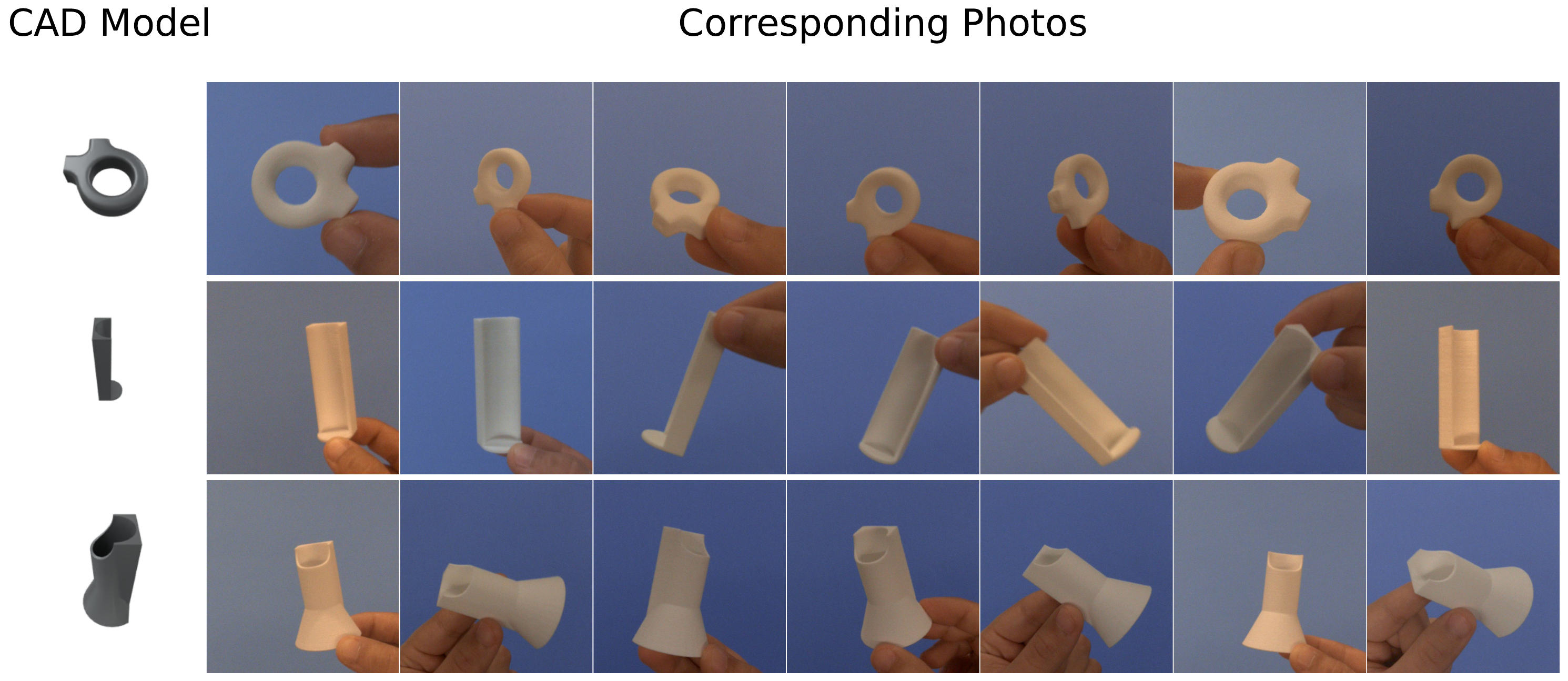}
    \caption{ThingiPrint Dataset. Examples of paired CAD models and real images of their 3D-printed counterparts. Our dataset includes 100 CAD models sourced from the Thingi10K dataset, each accompanied by multiple real-world photographs of the corresponding 3D-printed objects.}
    \label{fig:data_vis}
\end{figure}

Recent advances in representation learning have led to the emergence of general-purpose vision models~\cite{caron2021emerging,chen2020simple,grill2020bootstrap,he2020momentum,he2022masked, caron2020unsupervised} trained on large-scale natural image datasets. These models have demonstrated impressive transferability to various downstream tasks. However, they are often optimized for natural image domains and may not generalize well to the geometric and visual characteristics of CAD models and 3D printed objects. We argue that a vision model tailored to this domain could be significantly more effective for the classification task, as demonstrated by our experimental results (Table.~\ref{tab:model_accuracy_1}).

To facilitate research in this area, we introduce ThingiPrint, a new dataset of real photos of 3D-printed objects fabricated using CAD models drawn from the Thingi10K dataset  \cite{zhou2016thingi10k}. The dataset is intended as a benchmark for evaluating visual classification on novel 3D-printed objects. For a subset of these objects, we also provide an auxiliary version printed using a different 3D printer, enabling evaluation of printer-specific variations in classification performance and model robustness. We illustrate the dataset in Fig. \ref{fig:data_vis}.

To comprehensively assess the performance of current vision models on the task of classifying novel 3D-printed objects, we evaluate several pre-trained baseline models on our dataset. Given our goal of edge deployment, we opt for lightweight architectures that align with the computational constraints of our target setting. Beyond benchmarking existing models, we also propose a tailored contrastive training procedure. In particular, we fine-tune a pre-trained backbone on a large dataset of renderings of CAD models to adapt it to the 3D-printed object domain. Our method leverages a contrastive framework with a rotation-invariant objective. This encourages the model to map nearby viewpoints of the same object to similar feature representations, directly addressing the inherent viewpoint variability present in real-world photographs of 3D-printed objects. Our experimental results demonstrate the effectiveness of this method, highlighting the benefits of contrastive fine-tuning and view-invariant training in improving classification accuracy and robustness.

Our main contributions can be summarized as follows:
\begin{itemize}
    \item We introduce ThingiPrint, a dataset that pairs CAD models with real photographs of their 3D-printed counterparts. Additionally, we provide a smaller version of the dataset with objects printed using a different printer, designed to test printer-dependent generalization. The dataset will be made publicly available to support future research.
    \item We demonstrate that contrastive fine-tuning with a rotation-invariant objective produces a model capable of accurately classifying real images of unseen 3D-printed objects, outperforming other baselines.
\end{itemize}

\section{Related Work}
\label{sec:formatting}

\subsection{Vision in Additive Manufacturing}
Computer vision techniques have been adopted in a range of tasks related to additive manufacturing to support tasks such as object detection, recognition, and CAD generation. For example, Lemos et al.~\cite{lemos2019convolutional} proposed a convolutional neural network (CNN) for detecting 3D-printed objects to aid in automating the manufacturing process. More recently, CADCrafter~\cite{chen2025cadcrafter} introduced a deep learning framework that converts real-world images of manufactured objects into valid parametric CAD models. By leveraging synthetic renderings and geometry-aware encoders, such systems enable reverse engineering of 3D-printed parts directly from photographs. Focusing on object recognition, Conrad et al.~\cite{conrad2023recognition} developed a fully automated pipeline that uses synthetic CAD renderings to train a CNN capable of recognizing real-world 3D-printed parts. However, their approach requires retraining when encountering new objects, which limits adaptability in dynamic manufacturing settings. In contrast, our work focuses on 3D-printed object recognition with the goal of generalizing to previously unseen objects without the need for explicit retraining.

\subsection{CAD-to-Image Aligned Datasets}
Several existing datasets pair 3D models with real images. Pix3D~\cite{sun2018pix3d} provides instance-level CAD-to-photo alignments, primarily for furniture, where each CAD model aligns with a single real image. Pascal3D+~\cite{xiang2014beyond} and ObjectNet3D~\cite{xiang2016objectnet3d} offer category-level alignments, using representative CAD models for common object categories. ABO\cite{collins2022abo} extends this to consumer products, featuring high-quality 3D models alongside real product images. Furthermore, pose estimation datasets such as LINEMOD~\cite{hinterstoisser2012model}, YCB-Video~\cite{xiang1711posecnn}, and T-LESS~\cite{hodan2017t} contain real images of physical objects with known 3D models, often derived from industrial components or everyday items, and typically include only a limited number of objects.  In contrast to previous datasets, ThingiPrint focuses exclusively on 3D-printed objects derived from the Thingi10K database, making it specifically tailored to the 3D printing domain. Most related to our work, a recent study~\cite{chen2025cadcrafter} on CAD generation from real images introduced a dataset containing paired CAD models and real images of their 3D-printed counterparts, captured by placing the objects on a flat surface. However, this dataset is not publicly released, limiting its utility for benchmarking and further research. In contrast, we release our dataset publicly. Moreover, while their data contains only four images per object, we provide roughly ten images per object with greater viewpoint variation. Our objects are also handheld rather than placed on a surface, better reflecting our target use case of object recognition.

\section{Problem Formulation \& Dataset Construction}\label{data_collection}

\begin{figure*}[t]
    \centering
    \includegraphics[width=\textwidth]{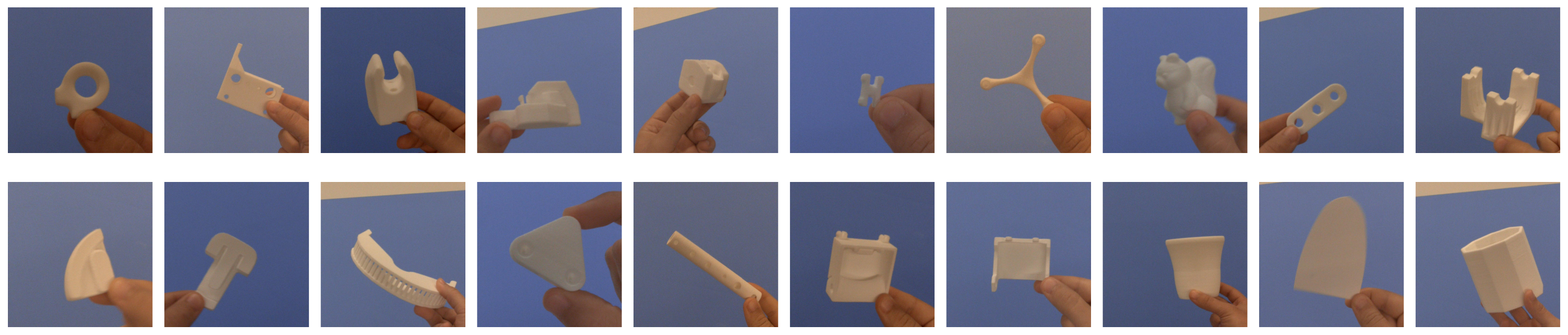}
    \caption{Examples of objects in ThingiPrint. Our dataset consists of various 3D-printed objects with diverse geometries.}
 \label{fig:objects}
\end{figure*}

\subsection{Problem Statement and Motivation}
We formalize the task as a prototype-based image classification problem for 3D-printed objects using synthetic renderings. Given a pre-trained encoder and a set of \( N \) CAD models \(\{M_1, M_2, \ldots, M_N\}\) with their corresponding synthetic renderings \(\{R_1, R_2, \ldots, R_N\}\), where \( R_i = \{r_i^1, r_i^2, \ldots, r_i^K\} \) represents \( K \) rendered views of model \( M_i \), the goal is to classify real photographs \(\{I_1, I_2, \ldots, I_T\}\) of these 3D-printed objects without training on the specific set of objects.

To evaluate models on this challenging task, we construct a test set that aligns CAD models with real images of their 3D-printed counterparts captured under varying orientations. Additionally, for a subset of objects, we include images of prints produced using a different printer and material to test generalization across varying printing conditions. This dataset design mirrors real-world deployment scenarios, where users capture images of 3D-printed objects under various conditions and viewpoints.

\subsection{Test Data Collection}

\textbf{Object Selection and 3D Printing.}
We randomly select 100 CAD models from the Thingi10K dataset \cite{zhou2016thingi10k}, a large collection of 3D printing models, to form our test set. Each selected CAD model was 3D-printed using an industrial-grade SLS printer (Sindoh S100) with white PA12 (polyamide 12) powder to maintain consistent appearance conditions across the primary dataset. See Fig.~\ref{fig:objects} for examples of objects contained in ThingiPrint.
\begin{figure}[h]
    \centering
    \includegraphics[width=\columnwidth]{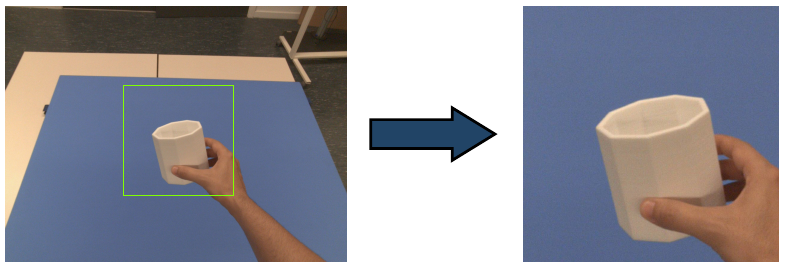}
    \caption{Example of the data collection process using smart glasses. The user positions the object within the rectangular region displayed on the device, and this region is cropped to obtain the final image.}
    \label{fig:crop}
\end{figure}

\begin{figure}[t]
    \centering
    \includegraphics[width=\columnwidth]{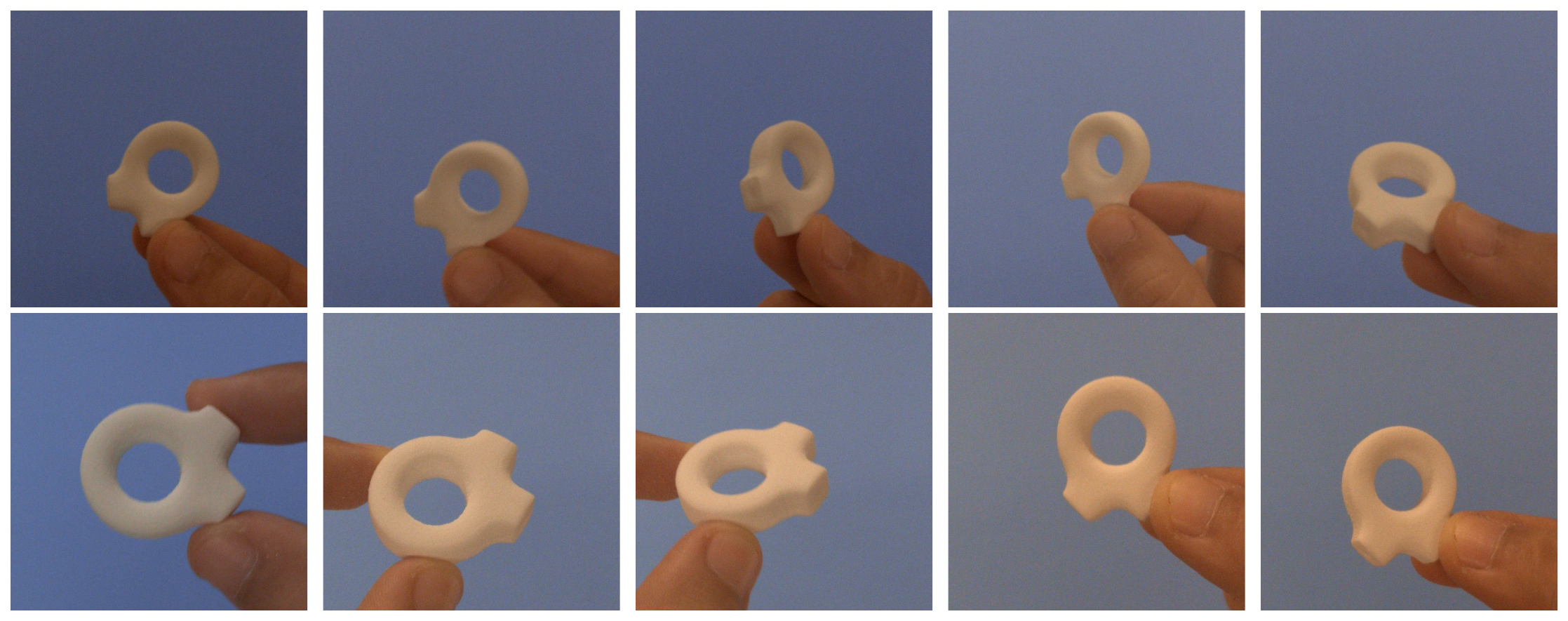}
    \caption{Object capture. Example photos of a single object from our dataset. The object is hand-held and rotated to capture a variety of orientations.}
    \label{fig:images_per_object}
\end{figure}

\textbf{Image Acquisition Protocol.}
For each printed object, we captured 10 photographs using a smart-glasses device while users naturally rotated objects by hand. This setup mimics realistic usage scenarios where objects can be viewed from different orientations during inspection. During data collection, a bounding box interface was displayed through the smart glasses to guide object positioning (see Fig.~\ref{fig:crop}). Upon capture, the region within the bounding box was automatically cropped, ensuring object-centric framing and eliminating the need for object localization. Angles that significantly obscured the object's distinguishing features were generally avoided, while natural manipulation patterns were maintained (see Fig.~\ref{fig:images_per_object}). This protocol ensures that the dataset reflects realistic operating conditions, allowing model evaluations to provide a reliable estimate of real-world performance.

\textbf{Cross-Printer Generalization Subset.}
To investigate model robustness under varying printing conditions, we selected 20 objects from the original set and reprinted them using a consumer-grade printer (Prusa MK4) with white PLA filament. This printer-material combination introduces variations in surface texture and reflectivity compared to the industrial setup. Following the identical image acquisition protocol, we captured 10 additional photographs per object, resulting in 200 images for cross-printer evaluation. This subset enables analysis of model generalization across varying physical manufacturing conditions.

\begin{figure*}[t]
    \centering
    \includegraphics[width=\textwidth]{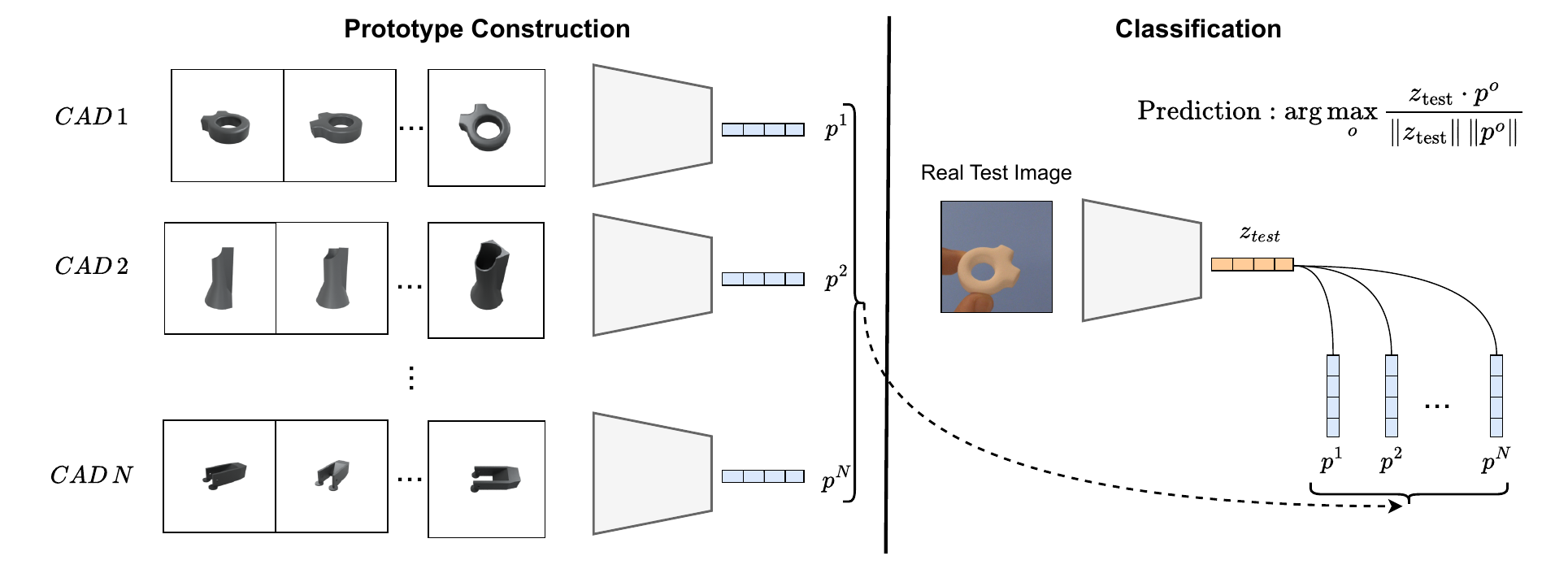}
    \caption{Overview of the classification pipeline at inference. (Left): At test time, we render images for each CAD model and compute a prototype vector $p_i$ to represent each object by averaging the representations of the encoded images. (Right): During inference, we classify an image of a 3D-printed object by comparing its feature representation $z_t$ to the computed prototype vectors and picking the object ID $o$ with the highest similarity.}
    \label{fig:method}
\end{figure*}

\subsection{Dataset Statistics and Composition}

The complete test set comprises 1,200 real photographs: 1,000 images from the primary industrial printing setup (100 objects, $\sim$ 10 images per object) and 200 images from the cross-printer subset (20 objects, $\sim$  10 images per object). Each object is also paired with multiple synthetic renderings of the corresponding CAD model, constituting the support set required for few-shot classification evaluation. The dataset encompasses a diverse set of objects, ensuring broad geometric variation for thorough model evaluation.

\section{Method}

\noindent\textbf{Overview.}
In this section, we describe our training and inference procedures. The training phase adapts the encoder to our task and is performed only once. After training, the encoder can be reused to classify any collection of objects. Inference consists of a prototype construction stage followed by the classification stage. During prototype construction, we compute a prototype vector for each object in the classification set. Once these prototypes are established, classification is carried out by comparing query features with the corresponding prototype vectors (see Fig.~\ref{fig:method}). When a new set of objects is introduced, the prototype construction stage must be repeated for the new set, while the trained encoder itself remains unchanged.

\subsection{Contrastive Training}
\label{sec:contrastive}

\begin{figure}[h]
    \centering
    \includegraphics[width=0.7\columnwidth]{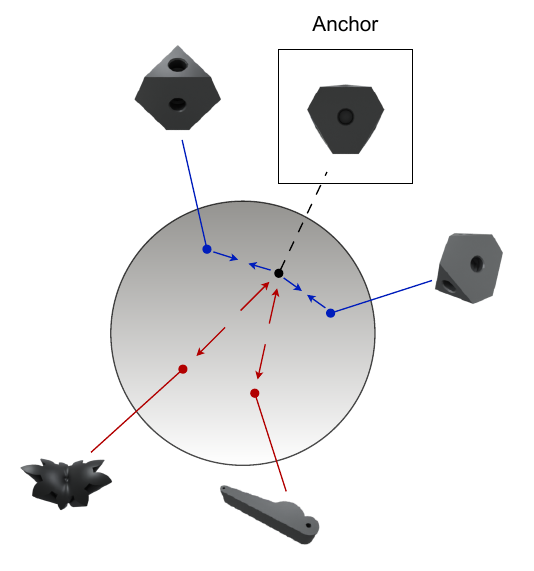}
    \caption{Rotation-invariant training. Different views of the same object are encouraged to have similar representations, while views of other objects serve as negative examples in the contrastive training framework.}
    \label{fig:rotation_invariance}
\end{figure}

To train a general-purpose feature extractor for novel 3D-printed object classification, we fine-tune our backbone network on a large set of synthetic images using a contrastive learning objective designed to encourage viewpoint-invariant representations. The training data consists of synthetic renderings from 6000 CAD models sourced from Thingi10K, which are disjoint from the 100 objects used in ThingiPrint. In addition to standard augmentations commonly used in contrastive learning, such as color jittering and random cropping, we introduce 3D-aware transformations by applying rotations in azimuth, elevation, and in-plane angles. For each training sample, a positive pair is constructed by rendering a nearby view of the same object, obtained by shifting the azimuth or elevation by 30 degrees and applying a random in-plane rotation. Negative samples are drawn from views of different objects (Fig.~\ref{fig:rotation_invariance}). To reduce the synthetic-to-real domain gap, we additionally apply background augmentation using real images collected in an industrial workspace, including hand-held interaction. We follow the standard contrastive loss, where given an anchor image \( x_i \), a positive sample \( x_i^+ \), and a set of negative samples \( \{x_j^-\}_{j=1}^{N} \), the objective encourages the similarity between \( x_i \) and \( x_i^+ \) to be higher than that between \( x_i \) and any \( x_j^- \). This is typically implemented using the InfoNCE loss:

\begin{equation}
\mathcal{L}_i =
- \log
\frac{
\exp(s_{i,i^+} / \tau)
}{
\sum_{k} \exp(s_{i,k} / \tau)
}
\label{eq:infonce}
\end{equation}

Here:
\begin{itemize}
    \item \( s_{i,k} = \text{sim}(f(x_i), f(x_k)) \), where \( k \) indexes the positive and all negative samples in the batch,
    \item \( f(\cdot) \) is the feature extractor network,
    \item \( \text{sim}(u, v) = \frac{u^\top v}{\|u\| \|v\|} \) is cosine similarity,
    \item \( \tau \) is a temperature hyperparameter.
\end{itemize}

The overall loss is computed as the average over all samples in the batch:

\begin{equation}
\mathcal{L} = \frac{1}{B} \sum_{i=1}^{B} \mathcal{L}_i
\end{equation}

\subsection{Inference}
To perform classification in an open-ended fashion, we adopt a prototype-based classification approach leveraging the available CAD models. For each object in the target set, we render images from a predefined set of viewpoints to capture appearance variations. These rendered images are then passed through the trained feature extractor to obtain a set of feature representations. 

Formally, for an object \( o \), let \( \{x_1^o, x_2^o, \dots, x_K^o\} \) denote the rendered images from \( K \) viewpoints. We extract their features using the encoder \( f(\cdot) \) as:
\begin{equation}
\{z_1^o, z_2^o, \dots, z_K^o\} = \{f(x_1^o), f(x_2^o), \dots, f(x_K^o)\}
\end{equation}

Subsequently, these features are aggregated into a single prototype vector for each object class by computing their mean:
\begin{equation}
p^o = \frac{1}{K} \sum_{k=1}^{K} z_k^o
\end{equation}
where \( p^o \) denotes the prototype representation for object \( o \), and \( \{z_k^o\}_{k=1}^{K} \) are the feature vectors extracted from \( K \) rendered views.

Given a test image \( x_{\text{test}} \), the prediction is obtained by computing the cosine similarity between its feature representation \(z_{\text{test}} =  f(x_{\text{test}}) \) and the prototype vector of each class, and selecting the class with the highest similarity score:
\begin{equation}
\hat{o} = \arg\max_{o} \, \frac{z_{\text{test}}^\top p^o}{\|z_{\text{test}}\| \, \|p^o\|}
\end{equation}

This formulation decouples the model from any fixed class set, enabling classification over arbitrary collections of objects. Each new classification set requires only the construction of new prototype vectors, with no need to retrain the model.

\section{Experiments}
We present the evaluation results of several models on our released dataset and compare their performance to our proposed fine-tuning strategy. Furthermore, we provide an in-depth quantitative analysis to explore how various factors—both during training and inference—affect the performance of our model.

\paragraph{Implementation Details.} For the training set we render images from all 6000 objects from two elevation angles (30\textdegree, 60\textdegree) and for 12 renderings per elevation for every 30 degrees in azimuth, resulting in a total of 144,000 images. For contrastive training, we use a batch of 350 objects (the most we can fit on a consumer-grade GPU) and use the AdamW \cite{loshchilov2017decoupled} optimizer with a learning rate of 1e-7. For rendering images for the construction of the prototypes at test time, we follow the same procedure as the training images, resulting in 24 images per object.

\paragraph{Models.} For our experimental comparisons, we select a set of pretrained vision models and compare them to our fine-tuned variants. For the pretrained models, we select models trained with different forms of supervision, allowing us to test how different pretraining objectives translate to downstream performance on our task. Specifically, we include an ImageNet-supervised ResNet-50~\cite{he2016deep} trained on labeled image classification; CLIP (ViT-B/32)~\cite{radford2021learning}, which employs contrastive language-image pretraining to learn multimodal representations; and DINOv2 (ViT-B/14)~\cite{oquab2023dinov2}, a self-supervised model trained via self-distillation, known for producing rich and transferable features. For our fine-tuned models, we start from the corresponding pretrained backbone and apply a fine-tuning stage on synthetic renderings generated from a separate corpus of 6,000 Thingi10K objects that is disjoint from the test objects in ThingiPrint.

\begin{table}[t]
\centering
\resizebox{\columnwidth}{!}{%
\begin{tabular}{lcc}
\toprule
\textbf{Model} & \textbf{Top-1 (\%)} & \textbf{Top-5 (\%)} \\
\midrule
\textbf{\textit{Pretrained Models}} & & \\
CLIP  & 27.1 & 61.0 \\
ResNet50 & 35.9 & 67.7 \\
DINOv2  & 61.8 & 86.4 \\
\midrule
\textbf{\textit{Fine-tuned Models}} & & \\
ResNet50  & 59.7\improvement{+23.8} & 84.7\improvement{+17} \\
DINOv2 & 76.5\improvement{+14.7} & 94.0\improvement{+7.6} \\
\bottomrule
\end{tabular}%
}
\caption{Comparison of pretrained and fine-tuned models on ThingiPrint. Subscript deltas indicate absolute improvements over the corresponding pretrained model.}
\label{tab:model_accuracy_1}
\end{table}

\subsection{3D-Printed Object Classification}
We use ThingiPrint to evaluate the performance of different vision models on the 3D-printed object classification task. We train on synthetic renderings of 6{,}000 objects from the Thingi10K dataset. Note that the objects of the training set are disjoint from the objects used in  ThingiPrint. The experimental results are presented in Table~\ref{tab:model_accuracy_1}. As shown, non-fine-tuned models struggle to achieve high classification accuracy on our test set, with CLIP and ResNet-50 reaching only 27.1\% and 35.9\%, respectively. DINOv2 performs better, achieving 61.8\% accuracy, reflecting its strong general feature extraction capabilities. Fine-tuning substantially improves performance for both backbones. The fine-tuned ResNet{-}50 reaches 59.7\% accuracy while the fine-tuned DINOv2 achieves the best overall performance with 76.5\% accuracy, a 14.7\% absolute gain over its pretrained counterpart. Furthermore, the fine-tuned DINOv2 attains a top-5 classification accuracy of 94.0\%, suggesting that the correct object is almost always retrieved among a small set of candidate predictions.

\begin{figure}[t]
    \centering
    \includegraphics[width=\columnwidth]{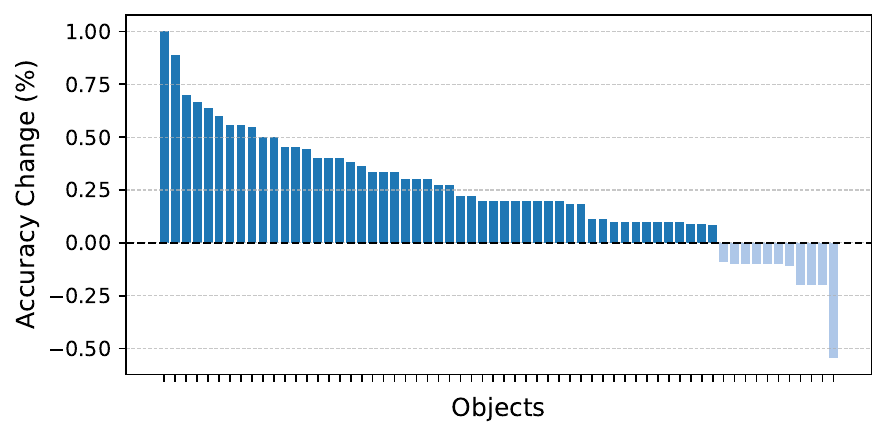}
    \caption{Per-object accuracy change. Fine-tuning leads to improved accuracy for most objects, though a few objects experience a decline in performance.}
    \label{fig:per_class}
\end{figure}

In Fig.~\ref{fig:per_class}, we present the per-object change in accuracy after fine-tuning, compared to the pretrained model. We report only the objects that exhibited a change in accuracy, whether positive or negative. As shown, fine-tuning generally leads to improved accuracy across many objects, though a few experience a decrease. Overall, these results indicate that our training consistently improves performance for the majority of objects.

\begin{figure}[t]
    \centering
    \includegraphics[width=1.0\columnwidth]{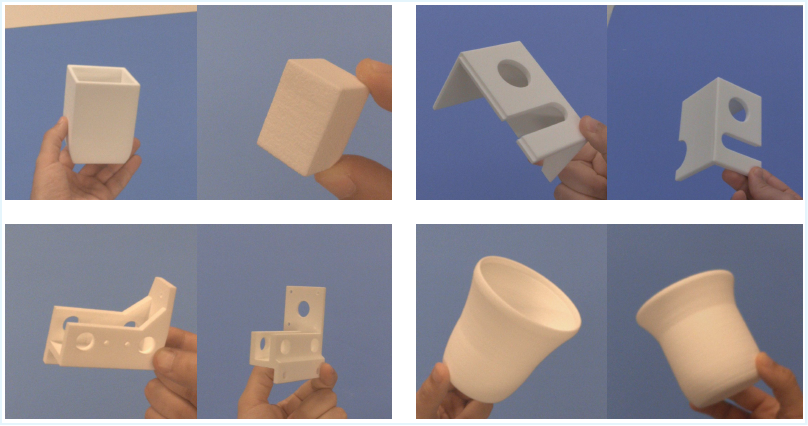}
    \caption{Visually similar objects. The dataset includes some objects with similar geometric shapes or symmetric counterparts, which pose challenges for the classification task.}
    \label{fig:similar_objects}
\end{figure}

\paragraph{Similar Objects.} The dataset contains objects with similar geometry or symmetric counterparts, presenting more challenging classification cases. Examples of these objects are shown in Fig.~\ref{fig:similar_objects}. To evaluate how challenging the classification task is for the model, we identified some visually similar object pairs in the test set. Using the model's embeddings, we computed the cosine similarity between class prototypes and ranked all class pairs accordingly. We then selected the top 20 most similar pairs, which corresponded to 18 unique objects. These were used to analyze the model's performance on the most confusing or visually similar categories. The classification accuracies of our model and other baselines for this subset are reported in Table~\ref{tab:accuracy_similar}. As expected, the overall accuracy for each model is lower on this subset compared to the original test set. Nevertheless, our fine-tuned model demonstrates significantly improved robustness, achieving an accuracy of 63.4\%, compared to the next best model, DINOv2, which achieves only 49.4\%. This indicates that our fine-tuned model learns more discriminative and robust feature representations for these challenging cases.

\begin{table}[h]
\centering
\resizebox{\columnwidth}{!}{%
\begin{tabular}{lcc}
\toprule
\textbf{Model} & \textbf{Top-1 (\%)} & \textbf{Top-5 (\%)} \\
\midrule
\textbf{\textit{Pretrained Models}} & & \\
CLIP & 16.2 & 54.2 \\
ResNet50 & 30.3 & 74.8 \\
DINOv2 & 49.4 & 92.2 \\
\midrule
\textbf{\textit{Fine-tuned Models}} & & \\
ResNet50  & 44.2\improvement{+13.9} & 88.5\improvement{+13.7} \\
DINOv2  & 63.4\improvement{+14.0} & 98.4\improvement{+6.2} \\
\bottomrule
\end{tabular}%
}
\caption{Classification accuracy on ThingiPrint for visually similar objects across models. Improvements in parentheses indicate gains over the corresponding pretrained model.}
\label{tab:accuracy_similar}
\end{table}

\subsection{Cross-Printer Generalization}
As discussed in \S\ref{data_collection}, we also construct a smaller dataset containing photographs of 20 objects from the original test set, reprinted using a different 3D printer, to assess whether printer-specific characteristics affect classification performance. The results of this comparison are presented in Table~\ref{tab:printer_comp}. Overall, we do not observe a substantial difference in classification accuracy between the two test sets. While the largest gap (5.8\%) occurs with the fine-tuned ResNet-50 model, this variation is likely due to random fluctuations on the small subset rather than systematic differences introduced by the printer. Notably, the other models — including both the pretrained and fine-tuned versions of DINOv2 — exhibit smaller differences. Since the geometry of the objects remains unchanged, variations in surface texture or light reflection introduced by different printers appear to have minimal impact. These results suggest that the printing method does not introduce a significant domain shift for this classification task.

\begin{figure}[t]
    \centering
    \includegraphics[width=0.8\columnwidth]{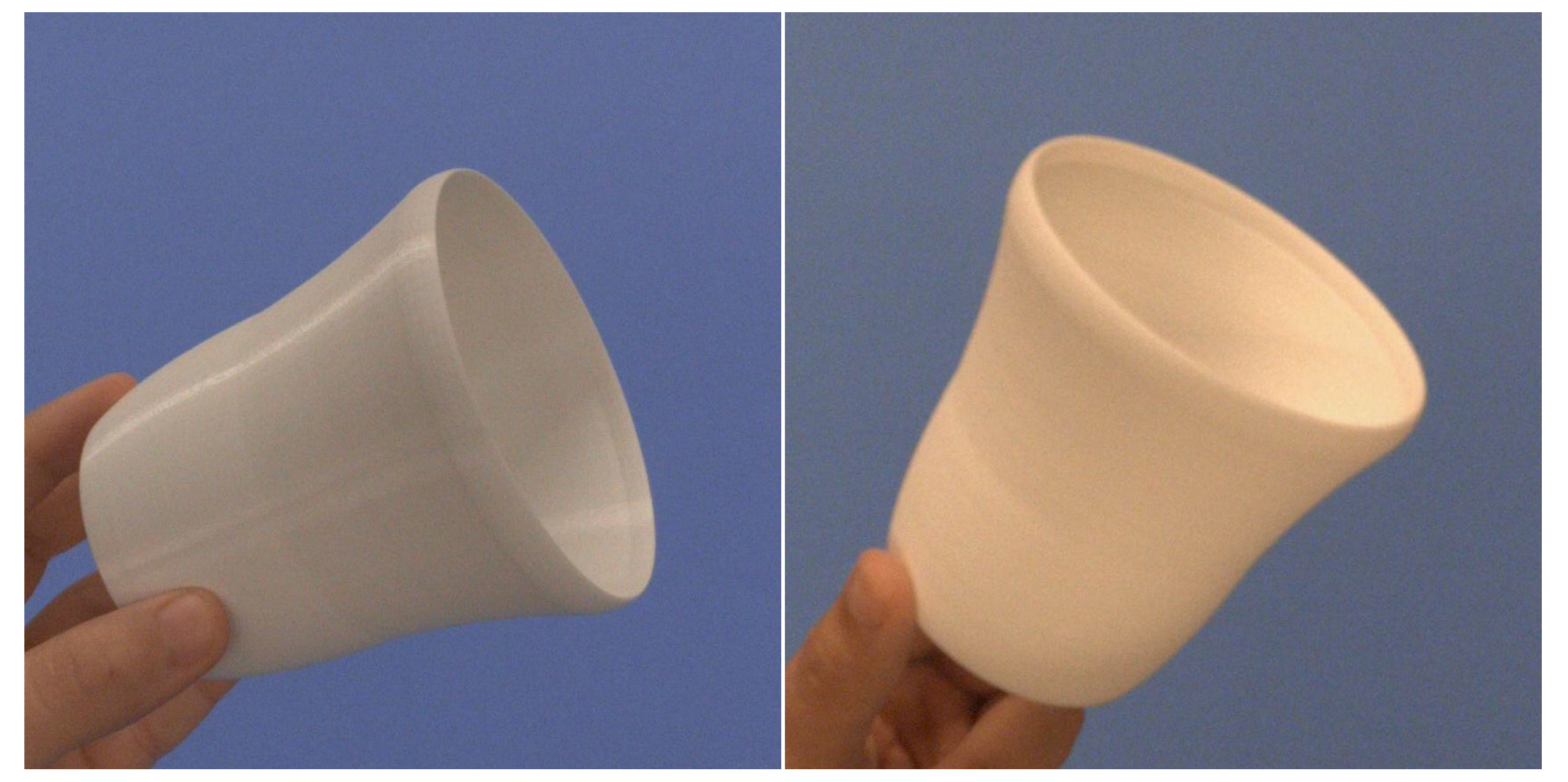}
    \caption{Printer-specific appearance. Example of the same object printed with different 3D printers. Left: printed with Prusa MK4. Right: printed with an industrial printer.}
    \label{fig:printers_matches}
\end{figure}

\begin{table}[t]
\centering
\resizebox{\columnwidth}{!}{%
\begin{tabular}{lclcc}
\toprule
              &                      & \multicolumn{2}{c}{\textbf{Printer type}} &  \\
\cmidrule(lr){3-4}
\textbf{Model} & \textbf{Synthetic Fine-Tuning} & \textbf{Industrial} & \textbf{Prusa MK4} & \textbf{$|\Delta|$}  \\
\midrule
ResNet50 & \xmark & 36.2 & 39.0 & 2.8 \\
ResNet50 & \cmark & 58.8 & 53.0 & 5.8 \\
\midrule
DINOv2 & \xmark & 69.5 & 72.9 & 3.4 \\
DINOv2 & \cmark & 80.9 & 79.0 & 1.9 \\
\bottomrule
\end{tabular}%
}
\caption{Top-1 classification accuracy of different models on the same 20-object subset printed with different 3D printers. The \textbf{$|\Delta|$} column represents the absolute difference of the accuracies between the two printers. }
\label{tab:printer_comp}
\end{table}

\subsection{Impact of Rotation Invariance}
Our training pipeline incorporates rotation invariance in contrastive learning, based on the assumption that rotation-invariant representations can enhance recognition performance. To qualitatively assess this, we conducted a systematic study on the effect of incorporating rotation invariance during contrastive training, with results summarized in.Table~\ref{tab:view_inv_effect}. As expected, the pretrained DINOv2 model achieves the lowest classification accuracy of 61.8\% when used directly for this task without fine-tuning. Applying standard contrastive fine-tuning on the Thingi dataset, using color jitter and crop augmentations, improves the accuracy to 68.9\%. Finally, incorporating a rotation-invariant contrastive objective as described in \S\ref{sec:contrastive} further boosts performance to 76.5\%. These results demonstrate that explicitly encouraging rotation-invariant representations is beneficial for this classification setting.

\begin{table}[t]
\centering
\resizebox{\columnwidth}{!}{%
\begin{tabular}{lccc}
\toprule
\textbf{Model} & \textbf{Synthetic Fine-Tuning} & \textbf{Rotation invariance} & \textbf{Top-1 (\%)} \\
\midrule
ResNet50 & \xmark    & \xmark    & 35.9 \\ 
ResNet50 & \cmark  & \xmark    & 56.2\improvement{+20.3} \\
ResNet50 & \cmark  & \cmark  & \textbf{59.7}\improvement{+23.8} \\
\midrule
DINOv2   & \xmark    & \xmark    & 61.8 \\
DINOv2   & \cmark  & \xmark    & 68.9\improvement{+7.1} \\
DINOv2   & \cmark  & \cmark  & \textbf{76.5}\improvement{+14.7} \\
\bottomrule
\end{tabular}%
}
\caption{Impact of rotation-invariance training on classification performance on ThingiPrint. Improvements are relative to the corresponding non-fine-tuned model.}
\label{tab:view_inv_effect}
\end{table}

\subsection{Test-Time Analysis}

Experiments in this section investigate how rendering configurations and multi-view aggregation at inference time influence the final classification performance. All results in this section are obtained using our fine-tuned DINOv2 model, which serves as the backbone for the classification task.

\begin{figure}[t]
    \centering
    \includegraphics[width=0.9\columnwidth]{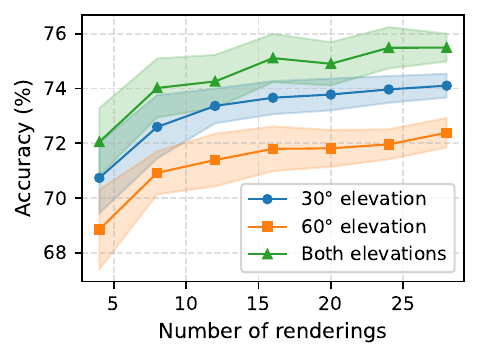}
    \caption{Number of renderings at inference. Effect of the number of rendered images on classification accuracy under different rendering strategies. We compare rendering images from a fixed elevation angle (30° or 60°) to rendering from both elevations while varying the azimuth angle. Increasing the number of renderings improves accuracy, with diminishing returns in all cases; combined elevation sampling consistently achieves the highest performance.}
    \label{fig:elev_angle}
\end{figure}

\paragraph{Effect of Viewpoint Diversity.} 
 To assess the effect of the rendering process on prototype construction at test time, we conducted a systematic analysis varying both the number of rendered images and their viewpoint distribution. As shown in Fig.~\ref{fig:elev_angle}, we plot classification accuracy as a function of the number of renderings under three viewpoint sampling strategies: (1) renderings from a fixed 30° elevation, (2) renderings from a fixed 60° elevation, and (3) a combination of both elevation angles. The first two curves correspond to the single-elevation conditions, where images are sampled by varying the azimuth angle. In both cases, increasing the number of renderings improves classification performance, with diminishing returns beyond 16 renderings. The third curve corresponds to sampling from both elevation angles, with renderings evenly split between them (e.g., for 8 renderings, 4 from each). This setup consistently outperforms the single-elevation baselines and achieves the highest accuracy, peaking around 24 renderings. These results highlight the benefits of viewpoint diversity in prototype construction, contributing to higher classification performance.

\begin{figure}[t]
    \centering
    \includegraphics[width=0.9\columnwidth]{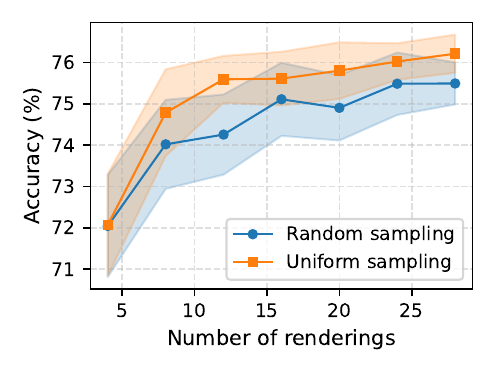}
    \caption{Viewpoint selection strategy. Comparison of random vs. uniform sampling strategies for rendering images around the object. The uniform sampling yields higher classification accuracy across different numbers of renderings, highlighting the advantage of better viewpoint coverage in prototype construction.}
    \label{fig:samling_method}
\end{figure}

\paragraph{Random vs. Uniform Viewpoint Sampling.} Building on the importance of viewpoint diversity, we further investigate how the sampling strategy itself affects prototype quality and classification performance. We compare two approaches: randomly sampling viewpoints around the object and using a uniform sampling strategy to maximize coverage. For these tests, we rendered images from both elevation angles, as described in the previous section. For each number of renderings, we repeated the evaluation 30 times and report the mean and standard deviation. As shown in Fig.~\ref{fig:samling_method}, the uniform sampling strategy consistently outperforms random sampling across all tested numbers of renderings. This demonstrates that uniform sampling provides better viewpoint coverage, leading to more robust prototype construction and improved classification accuracy.

\begin{figure}[t]
    \centering
    \includegraphics[width=0.9\columnwidth]{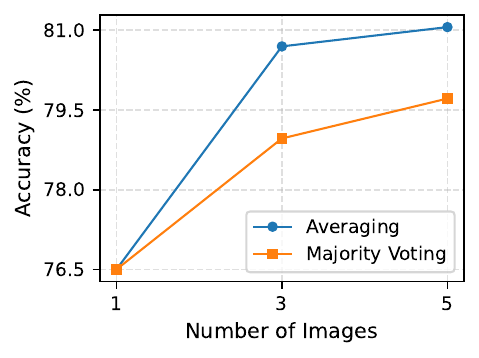}
    \caption{ Impact of aggregating predictions from multiple images at inference time on classification accuracy. Accuracy consistently improves with the number of views, demonstrating the benefit of multi-view prediction aggregation.}
    \label{fig:inference_views_aggr}
\end{figure}

\paragraph{Inference with Multi-View Aggregation.} Aggregating predictions from multiple images of a single object can improve robustness and accuracy by mitigating the impact of ambiguous or challenging viewpoints. We investigate how leveraging multiple views at inference time influences final classification performance by testing two aggregation methods: majority voting and averaging of predicted logits across images. In Fig.~\ref{fig:inference_views_aggr}, we present quantitative results showing the effect of incorporating multiple views during inference for an increasing number of images. As illustrated, classification accuracy consistently improves for both methods as the number of views increases, underscoring the advantages of multi-view aggregation. Averaging results in slightly higher accuracy than majority voting for each number of references. These results suggest that capturing a short video of the object or supplying multiple images at test time can  enhance classification robustness and accuracy, an important consideration for practical deployment.

\section{Conclusion}
We introduced ThingiPrint, a novel dataset that directly links CAD models with real photographs of their 3D-printed counterparts. To construct this dataset, we 3D-printed 100 diverse objects from the Thingi10K dataset and captured multiple images of each. This dataset serves as a valuable resource for evaluating vision models in the domain of 3D-printed object classification.

Using this dataset, we established baseline evaluations for various pretrained vision models and found that they often struggle with the task. To address this challenge, we explored contrastive fine-tuning with a view-invariant objective, which led to significant improvements in classification performance and greater robustness to visually similar objects. Finally, we conducted a comprehensive analysis of key components of our pipeline, offering insights that can guide practical use and improve test-time robustness.

\paragraph{Limitations.} ThingiPrint is constructed by taking photos of 3D printed objects placed against clean, uncluttered, and uniformly colored backgrounds. This setup simplifies object recognition, as the object is easily distinguishable from the background. While collecting data in more diverse environments with clutter and varying backgrounds could introduce additional challenges and better simulate unconstrained scenarios, such complexity is not essential for our intended use case. In fact, it may make the task artificially difficult without offering practical benefit. In realistic industrial or workshop settings, it is straightforward to standardize the imaging setup by using uniform backgrounds. Thus, although our dataset lacks background variation, this controlled setting reflects practical deployment scenarios and remains aligned with real-world usage.

\medskip
 \noindent\textbf{Funding}\\
 This research has been funded through research project C-DATA: ‘Digitalizing AM through data sharing and AI’ (HBC.2022.0135), funded by VLAIO (Flemish government agency Innovation \& Entrepreneurship).


\bibliography{sn-bibliography}


\end{document}